\documentclass[letterpaper, 10 pt, conference]{ieeeconf}  

\IEEEoverridecommandlockouts                              

\usepackage{hyperref}
\usepackage{url}
\usepackage{graphics} 
\usepackage{times} 
\usepackage{booktabs}       
\usepackage{amsfonts}       
\usepackage{nicefrac}       
\usepackage{microtype}      
\usepackage{algorithm,multirow,xcolor}
\usepackage{algorithmicx}
\usepackage{algpseudocode}
\usepackage{mathtools}
\usepackage{graphicx}
\usepackage{cases}
\usepackage{array}
\usepackage{color}
\usepackage{float}
\usepackage{amssymb}
\usepackage{cite}
\usepackage{bm}

\usepackage{wrapfig}
\usepackage{subfigure}
\usepackage{amsmath}
\usepackage{amsthm, amssymb}
\theoremstyle{definition}

\usepackage{soul}

\usepackage{graphicx} 
\usepackage{subfigure}
\usepackage{epstopdf}
\usepackage{cite}
\usepackage{enumerate}
\usepackage[skip=5pt]{caption}

\newtheorem*{remark}{Remark}

\title{\LARGE \bf

A Recursive Total Least Squares Solution for Bearing-Only Target Motion Analysis and Circumnavigation
}

\author{Lin Li$^{*}$\textsuperscript{\dag}, Xueming Liu$^{*}$\textsuperscript{\dag}, Zhoujingzi Qiu\textsuperscript{\ddag}, Tianjiang Hu\textsuperscript{\dag}, and Qingrui Zhang\textsuperscript{\dag}%
\thanks{This work is supported by the National Natural Science Foundation of China under Grant 62203480, 62473390, and in part by the Basic and Applied Basic Research Foundation of Guangdong Province under Grant 2024A1515012408. $^{*}$These authors contribute equally to this work. Correspondence to Qingrui Zhang (e-mail: zhangqr9@mail.sysu.edu.cn). 
Video available at https://b23.tv/DlLHFIb}
\thanks{\textsuperscript{\dag}School of Aeronautics and Astronautics, Shenzhen campus of Sun Yat-sen University, Shenzhen 518107,  China.} 
\thanks{\textsuperscript{\ddag}Shenzhen Institute for Advanced Study, University of Electronic Science and Technology of China, Shenzhen 518110, China.}%
}
\begin{document}

\maketitle
\thispagestyle{empty}
\pagestyle{empty}

\begin{abstract}
Bearing-only Target Motion Analysis (TMA) is a promising technique for passive tracking in various applications as a bearing angle is easy to measure. Despite its advantages, bearing-only TMA is challenging due to the nonlinearity of the bearing measurement model and the lack of range information, which impairs observability and estimator convergence. This paper addresses these issues by proposing a Recursive Total Least Squares (RTLS) method for online target localization and tracking using mobile observers. The RTLS approach, inspired by previous results on Total Least Squares (TLS), mitigates biases in position estimation and improves computational efficiency compared to pseudo-linear Kalman filter (PLKF) methods. Additionally, we propose a circumnavigation controller to enhance system observability and estimator convergence by guiding the mobile observer in orbit around the target. Extensive simulations and experiments are performed to demonstrate the effectiveness and robustness of the proposed method.  The proposed algorithm is also compared with the state-of-the-art approaches, which confirms its superior performance in terms of both accuracy and stability. 

\end{abstract}

\section{INTRODUCTION}


Bearing-only Target Motion Analysis (TMA) has garnered significant attention due to its passive sensing capabilities, offering notable advantages in applications such as marine biological tracking \cite{shinzakiMultiAUVSystemCooperative2013}, ground target surveillance \cite{deghatLocalizationCircumnavigationSlowly2014,liuFormationControlMoving2023,suiAdaptiveBearingOnlyTarget2024}, and aerial pursuit operations \cite{liThreeDimensionalBearingOnlyTarget2022,ningBearingangleApproachUnknown2024a}. Furthermore, recent advancements in sensor technologies, such as infrared, sonar, and monocular cameras, have significantly improved the cost-effectiveness and accuracy of bearing acquisition. However, bearing-only TMA encounters inherent challenges due to the nonlinearity of the bearing measurement model and the observability limitations resulting from the lack of range information.

To address the nonlinearity of bearing measurements, the pseudo-linear Kalman filter (PLKF) \cite{lingrenPositionVelocityEstimation1978, liThreeDimensionalBearingOnlyTarget2022, ningComparisonDifferentPseudolinear2023, ningBearingangleApproachUnknown2024a} has attracted considerable attention. By exploiting spatial geometric relationships, the PLKF establishes a pseudo-linear function that relates target states to bearing measurements, encapsulating nonlinear components within the noise term. Compared to traditional Extended Kalman Filter (EKF) methods, the PLKF offers enhanced computational efficiency and stability \cite{liThreeDimensionalBearingOnlyTarget2022}. However, this pseudo-linear transformation introduces non-Gaussian noise characteristics that depend on the true target-observer distance. Since the true distance is typically unavailable, it is often approximated by the estimated distance between the target and the observer \cite{liThreeDimensionalBearingOnlyTarget2022}. This substitution inevitably introduces bias in the position estimate \cite{hoAsymptoticallyUnbiasedEstimator2006}, although existing theory shows that the velocity estimate remains asymptotically unbiased \cite{aidalaBiasedEstimationProperties1982}.

Thus, mitigating position estimation bias has become a central focus in bearing-only TMA research. Several studies indicate that pseudo-linear transformations lead to a noisy measurement matrix, framing the problem as an error-in-variable (EIV) challenge \cite{hoAsymptoticallyUnbiasedEstimator2006,dogancayBearingsonlyTargetLocalization2005,guNovelPowerBearingApproach2011}. This has prompted the adoption of Total Least Squares (TLS) methods  \cite{hoAsymptoticallyUnbiasedEstimator2006,vaghefiBearingonlyTargetLocalization2010,guNovelPowerBearingApproach2011,crassidisErrorCovarianceAnalysisTotal2014,xuRobustConstrainedTotal2023,dogancayBearingsonlyTargetLocalization2005,dogancayRelationshipGeometricTranslations2008,farinaTargetTrackingBearings1999}, which extend traditional least-squares methods by accounting for noise in both measurement vectors and measurement matrices \cite{golubAnalysisTotalLeast1980}. Theoretical analyses and simulation studies have shown that TLS estimators asymptotically approach the Cramér-Rao Lower Bound (CRLB) and rival Maximum Likelihood (ML) estimators \cite{hoAsymptoticallyUnbiasedEstimator2006,dogancayBearingsonlyTargetLocalization2005,guNovelPowerBearingApproach2011,crassidisErrorCovarianceAnalysisTotal2014}. Moreover, TLS methods have shown robustness against uncertainties in observer position \cite{vaghefiBearingonlyTargetLocalization2010,dogancayBearingsonlyTargetLocalization2005,guNovelPowerBearingApproach2011,crassidisErrorCovarianceAnalysisTotal2014}. However, the computational complexity of TLS methods limits their applicability for real-time mobile observers. To address this challenge, this paper applies the Recursive Total Least Squares (RTLS) method \cite{rhodeRecursiveGeneralizedTotal2014, koideRobustRegularizedAlgorithm2024} to bearing-only TMA, enabling online target localization and tracking by a moving observer.
    \begin{figure}[tbp]
        \centering
        \includegraphics[width=0.8\linewidth]{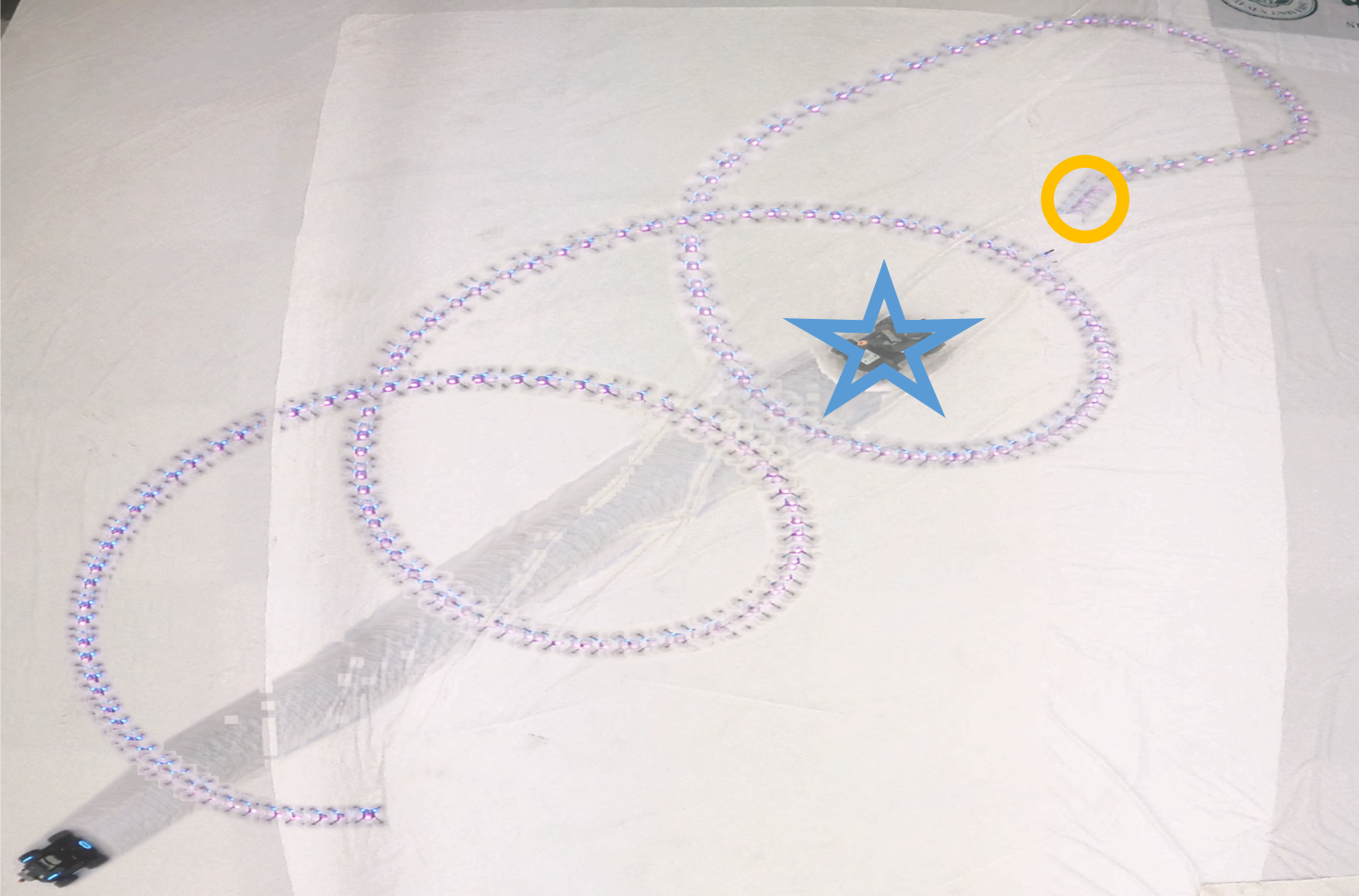}
        \caption{A drone circumnavigates a ground vehicle with the proposed bearings-only target motion analysis algorithm,the star symbol indicates the initial position of the target robot,the orange circle denotes the initial position of the UAV observer. }
        \label{fig:UAV-TARGET}
    \end{figure}
    
In addition, observability remains another fundamental challenge in bearings-only TMA due to the lack of distance information. Recent theoretical analyses using Fisher Information Matrix (FIM) \cite{liThreeDimensionalBearingOnlyTarget2022} have demonstrated that circumnavigation strategies significantly enhance system observability, ensuring estimator convergence and improving positioning accuracy. Building on these theoretical foundations, we propose a circumnavigation motion controller that guides the observer in orbiting the moving target using real-time state estimates, thus extending the framework established in \cite{deghatLocalizationCircumnavigationSlowly2014}.

In conclusion, the main contributions of this work are summarized as follows:
\begin{enumerate}
    
    \item An RTLS method, inspired by the works of \cite{rhodeRecursiveGeneralizedTotal2014} and \cite{koideRobustRegularizedAlgorithm2024}, is employed for bearings-only TMA with mobile observers. Compared to the state-of-the-art PLKF methods in \cite{liThreeDimensionalBearingOnlyTarget2022}, this approach significantly reduces bias in position estimation. Additionally, the impact of observer positional errors is explicitly considered in this study.

    \item A circumnavigation controller with bounded input is designed for the mobile observer by integrating the estimated target’s state, ensuring the convergence of RTLS.

    \item Extensive simulations and experiments were conducted, and the results demonstrate the accuracy improvement and robustness of the proposed method compared to existing technologies.  

\end{enumerate}

\textbf{Notations:}	
In this paper, $\mathbb{N}$ and $\mathbb{R}$ represent the sets of natural numbers and real numbers, respectively. The transpose of a matrix is indicated by the superscript ``$T$''. Let $\|\cdot\|$ denote the Euclidean norm of a vector, and $\|\cdot\|_F$ the Frobenius norm of a matrix. The identity matrix is denoted by $\bm{I}$, and $\bm{0}$ signifies the zero vector.

\section{Problem Statement}
\begin{figure}[tbp]
        \centering
        \includegraphics[width=0.85\linewidth]{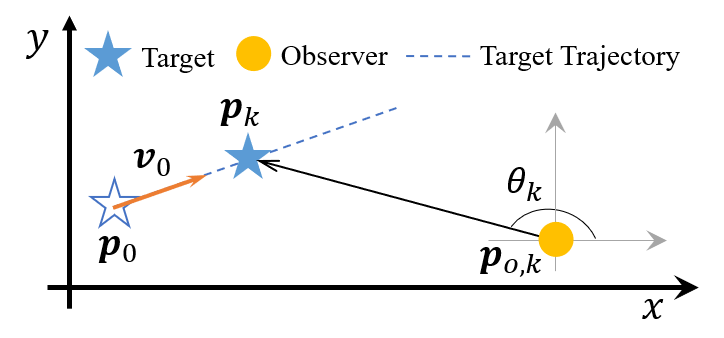}
        \caption{ Illustration of bearings-only target motion analysis. }
        \label{fig:frames}
    \end{figure}

In this section, the TMA and circumnavigation problem are formulated. For a better introduction to the algorithm, it is assumed that the target of interest moves with constant velocity along a straight-line trajectory in a two-dimensional plane, as shown in Fig.~\ref{fig:frames}. In this case, it is possible to characterize the kinematics of the target using the initial position $\boldsymbol{p}_0 \in \mathbb{R}^2$ and the initial velocity $\boldsymbol{v}_0 \in \mathbb{R}^2$, which correspond to the core TMA parameters \cite{guNovelPowerBearingApproach2011,hoAsymptoticallyUnbiasedEstimator2006,streitLinearLeastSquares1999}. Let $\boldsymbol{p}_k\in\mathbb{R}^2$ be the target position at time $k \Delta t$ and $\boldsymbol{v}_k\in\mathbb{R}^2$ be the velocity also at time $k \Delta t$. Hence, one has
    \begin{equation}\label{eq:target_dyn}
        \boldsymbol{p}_k = \boldsymbol{p}_0 + k \Delta t \boldsymbol{v}_0, \quad \boldsymbol{v}_k = \boldsymbol{v}_0
    \end{equation}
where $\boldsymbol{p}_k \triangleq \left[p^x_{k}, p^y_{k}\right]^T$, $k \in \mathbb{N}$ denotes the discrete time index and $\Delta t$ represents the sampling time interval.

The TMA process is conducted by a mobile observer (\emph{e.g.} a quadrotor) that performs real-time localization and tracking of a moving target using bearing-only measurements. The mobile observer's dynamics follow a first-order discrete-time model with bounded control input
    \begin{equation}\label{eq:uav_dyn}
        \boldsymbol{p}_{o,k+1} = \boldsymbol{p}_{o,k} + \Delta t \boldsymbol{u}_{o,k}, \quad \|\boldsymbol{u}_{o,k}\| \leq U
    \end{equation}
where $\boldsymbol{p}_{o,k} \triangleq \left[p^x_{o,k}, p^y_{o,k}\right]^T  \in \mathbb{R}^2$ represents the mobile observer's position vector, and $\boldsymbol{u}_{o,k} \in \mathbb{R}^2$ denotes the control input with a upper bound $U$.

It is assumed that the mobile observer only has access to bearing angle measurements $\tilde{\theta}_{k}$ relative to the motion target at each time step $k$. The measurement model is, therefore, expressed as
    \begin{equation}\label{eq:bearing}
        \tilde{\theta}_{k} = \theta_{k} + \mu_{\theta,k}, \quad \theta_{k} = \tan^{-1} \frac{p^y_k - p^y_{o,k}}{p^x_k - p^x_{o,k}}
    \end{equation}
where $\theta_{k}$ represents the true bearing angle, $\mu_{\theta,k} \backsim \mathcal{N}(0,\sigma_{\theta}^2)$ denotes zero-mean Gaussian measurement noise with variance $\sigma_{\theta}^2$. It is assumed that a mobile observer needs to conduct self-localization, which is a common case in severe applications, \emph{e.g.}, in GPS-denied environments.  The mobile observer's self-localization is commonly subject to measurement errors, so one has
    \begin{equation}\label{eq:observer_pos}
        \tilde{\boldsymbol{p}}_{o,k} = \boldsymbol{p}_{o,k} + \boldsymbol{\mu}_{\boldsymbol{p},k}, \quad \boldsymbol{\mu}_{\boldsymbol{p},k} \backsim \mathcal{N} \left( \bm{0},\sigma_{p}^2 \bm{I} \right)
    \end{equation}
where $\tilde{\boldsymbol{p}}_{o,k} \triangleq \left[\tilde{p}^x_{k}, \tilde{p}^y_{k}\right]^T$ represents the measured position vector, and $\boldsymbol{\mu}_{\boldsymbol{p},k} \triangleq \left[\mu^x_{k}, \mu^y_{k}\right]^T$ denotes an independent bivariate Gaussian noise with zero mean and diagonal covariance matrix $\sigma_{p}^2 \bm{I}$.

It should be noted that there are always errors in a mobile observer's self-localization, even in the case where GPS signals are available. Hence, Eq. \eqref{eq:observer_pos} is a common measurement model.

In this paper, we aim to find a robust and reliable solution to persistently estimate the motion of a target with a mobile observer. Hence, there are two sub-tasks:
\begin{enumerate}
    \item Developing a robust estimation framework to estimate a moving target's position and velocity using noisy bearing measurements and imperfect mobile observer self-localization data.
    \item Designing a circumnavigation control strategy that can generate feasible motion for a mobile observer to ensure estimator convergence and maintain stable target tracking performance.
\end{enumerate}

\section{measurement models}
According to the geometric relationship between a target and a mobile observer, the bearing angle $\theta_k$ satisfies the following trigonometric equation.
    \begin{equation}\label{eq:tan}
        \tan\theta_k=\frac{\sin\theta_k}{\cos\theta_k}=\frac{p^y_k - p^y_{o,k}}{p^x_k - p^x_{o,k}}
    \end{equation}
where both measurement noises and self-localization errors are ignored at this stage. Through a straightforward mathematical transformation of \eqref{eq:tan}, it is easy to derive
    \begin{equation} \label{eq:linear_true}
        \begin{bmatrix} \sin\theta_k & -\cos\theta_k \end{bmatrix} \boldsymbol{p}_{o,k} = \begin{bmatrix} \sin\theta_k & -\cos\theta_k \end{bmatrix} \boldsymbol{p}_{k}
    \end{equation}
Following \eqref{eq:target_dyn}, the TMA parameter, which corresponds to the initial conditions of the moving target, is defined as $\boldsymbol{x} = \begin{bmatrix} \boldsymbol{p}_0^T & \boldsymbol{v}_0^T \end{bmatrix}^T$, so one has
    \begin{equation} \label{eq:TMA_para}
        \boldsymbol{p}_{k} = \bm{M}_k \boldsymbol{x}, \quad \bm{M}_k \triangleq \begin{bmatrix} \bm{I} & k \Delta t \bm{I} \end{bmatrix}
    \end{equation}
Here,$\bm{I}$ denotes the identity matrix.Substituting \eqref{eq:TMA_para} into \eqref{eq:linear_true} results in the linear measurement model as given in \eqref{eq:linear_measurement}.
    \begin{equation} \label{eq:linear_measurement}
        y_k = \boldsymbol{h}_k^T \boldsymbol{x}
    \end{equation}
where 
\begin{align*}
    y_k &= \begin{bmatrix} \sin\theta_k & -\cos\theta_k \end{bmatrix} \boldsymbol{p}_{o,k} \\
    \boldsymbol{h}_k^T &= \begin{bmatrix} \sin\theta_k & -\cos\theta_k \end{bmatrix} \bm{M}_k
\end{align*}

Note that the measurement model \eqref{eq:linear_measurement} is derived in an ideal condition, namely ignoring the existence of measurement errors. However, in real applications, there are also noises or biases in measurements as given in \eqref{eq:bearing} and \eqref{eq:observer_pos}. Hence,  with the consideration of measurement noises, the ideal values in \eqref{eq:linear_measurement} should be replaced by their true counterparts, leading to a nonlinear scenario where errors show up in sinusoidal and cosinusoidal functions. It is necessary to apply a first-order approximation \cite{crassidisErrorCovarianceAnalysisTotal2014},  \emph{i.e.} $\sin \tilde{\theta}_k = \sin \theta_k + \mu_{\theta,k} \cos \theta$ and $\cos \tilde{\theta}_k = \cos \theta_k - \mu_{\theta,k} \sin \theta$. One, therefore, has 
    \begin{equation} \label{eq:y_noise}
        \begin{aligned}
            \tilde{y}_k = \begin{bmatrix} \sin \tilde{\theta}_k & -\cos\tilde{\theta}_k \end{bmatrix} \tilde{\boldsymbol{p}}_{o,k} = y_k + \delta y_k 
        \end{aligned}
    \end{equation}
where
\begin{equation*}
    \delta y_k = (\cos \theta_k \tilde{p}^x_{k} + \sin \theta_k \tilde{p}^y_{k})\mu_{\theta,k} + \mu^x_k \sin \theta_k  - \mu^y_k \cos \theta
\end{equation*}
and
    \begin{equation} \label{eq:h_noise}
        \begin{aligned}
            \tilde{\boldsymbol{h}}_k^T =  \begin{bmatrix} \sin\tilde{\theta}_k & -\cos\tilde{\theta}_k \end{bmatrix} \bm{M}_k = \boldsymbol{h}_k^T + \delta\boldsymbol{h}_k^T  
        \end{aligned}
    \end{equation}
where
$$\delta\boldsymbol{h}_k^T = \begin{bmatrix} \mu_{\theta,k} \cos\theta_k  & \mu_{\theta,k} \sin\theta_k \end{bmatrix} \bm{M}_k $$
Both $\delta y_k$ and $\delta\boldsymbol{h}_k$ are characterized by zero-mean Gaussian random variables. Their covariance matrices are respectively given by
    \begin{equation} \label{eq:y_cov}
            r_{y,k} = \left(\cos \theta_k \tilde{p}^x_{k} + \sin \theta_k \tilde{p}^y_{k} \right)^2 \sigma_{\theta}^2 +\sigma_{p}^2
    \end{equation}
    \begin{equation} \label{eq:h_cov}
            \bm{R}_{h,k} = \bm{M}_k^T \begin{bmatrix} \cos^2 \theta_k & \sin\theta_k \cos\theta_k \\ \sin\theta_k \cos\theta_k & \sin^2 \theta_k \end{bmatrix} \bm{M}_k
    \end{equation}
Eventually, the following measurement model is obtained, which has better matches with real-world scenarios.
    \begin{equation} \label{eq:plinear_measurement}
        \tilde{y}_k - \delta y_k = \left( \tilde{\boldsymbol{h}}_k^T - \delta\boldsymbol{h}_k^T \right) \boldsymbol{x}
    \end{equation}

\begin{remark}
    The key distinction between the proposed measurement model \eqref{eq:plinear_measurement} and conventional pseudo-linear approaches \cite{lingrenPositionVelocityEstimation1978,liThreeDimensionalBearingOnlyTarget2022} lies in the independence of measurement noise from the true target-observer distance \cite{crassidisErrorCovarianceAnalysisTotal2014,dogancayBearingsonlyTargetLocalization2005}. While existing methods \cite{liThreeDimensionalBearingOnlyTarget2022} replace the true distance with its estimate, this substitution introduces two critical limitations. First, the noise covariance matrix may be overestimated during the estimator's convergence phase, leading to degraded convergence rates. Second, errors in the mobile observer's self-localization propagate into the distance estimation, introducing additional bias and further degrading estimation accuracy. In our measurement model, although the true value $\theta_k$ in \eqref{eq:y_cov} and \eqref{eq:h_cov} is unavailable and should be replaced by $\tilde{\theta}_k$, the resulting error remains negligible due to the first-order approximation and bounded nature of trigonometric functions.
\end{remark}

\section{Recursive Total Least Squares Algorithm}
Based on the measurement model in \eqref{eq:plinear_measurement}, the TMA parameters $\boldsymbol{x}$ can be estimated by solving the following EIV model
    \begin{equation} \label{eq:eiv}
        \tilde{\boldsymbol{y}} \approx \tilde{\bm{H}} \boldsymbol{x}, \quad \tilde{\boldsymbol{y}} =\boldsymbol{y} + \Delta \boldsymbol{y}, \quad \tilde{\bm{H}} = \bm{H} + \Delta \bm{H}
    \end{equation}
where
$$\tilde{\boldsymbol{y}} = \begin{bmatrix} \tilde{y}_0 \\ \cdots \\ \tilde{y}_N \end{bmatrix},  \boldsymbol{y} = \begin{bmatrix} y_0 \\ \cdots \\ y_N \end{bmatrix}, \Delta \boldsymbol{y} = \begin{bmatrix} \delta y_0 \\ \cdots \\ \delta y_N \end{bmatrix}$$
$$\tilde{\bm{H}} = \begin{bmatrix} \tilde{\boldsymbol{h}}_0^T \\ \cdots \\ \tilde{\boldsymbol{h}}_N^T \end{bmatrix},  \bm{H} = \begin{bmatrix} \boldsymbol{h}_0^T \\ \cdots \\ \boldsymbol{h}_N^T \end{bmatrix}, \Delta \bm{H} = \begin{bmatrix} \delta \boldsymbol{h}_0^T \\ \cdots \\ \delta \boldsymbol{h}_N^T \end{bmatrix}$$
To solve \eqref{eq:eiv}, we first introduce the TLS framework \cite{golubAnalysisTotalLeast1980}, followed by a recursive implementation \cite{rhodeRecursiveGeneralizedTotal2014, koideRobustRegularizedAlgorithm2024}.

\subsection{Total Least Squares}
The fundamental principle of TLS is to concurrently minimize the noise effects in both $\tilde{\bm{H}}$ and $\tilde{\boldsymbol{y}}$ under the Frobenius norm. The generalized TLS problem can be formulated as follows \cite{rhodeRecursiveGeneralizedTotal2014,golubAnalysisTotalLeast1980}
    \begin{equation} \label{eq:GTLS}
    \begin{array}{c}
         \min_{\boldsymbol{x},\hat{\bm{Z}}} \left\lVert \bm{\Lambda} \left( \tilde{\bm{Z}}-\hat{\bm{Z}} \right) \bm{W} \right \rVert_F \\
         \text{subject to  } \hat{\bm{Z}} \begin{bmatrix} \boldsymbol{x} \\ -1 \end{bmatrix} = \bm{0}
    \end{array}
    \end{equation}
where $\tilde{\bm{Z}} = \begin{bmatrix} \tilde{\bm{H}} & \tilde{\boldsymbol{y}} \end{bmatrix}$ denotes the augmented measurement matrix, and $\hat{\bm{Z}} = \begin{bmatrix} \hat{\bm{H}} & \hat{\boldsymbol{y}} \end{bmatrix}$ represents its noise-corrected counterpart. The solution $\hat{\boldsymbol{x}}$, obtained from the corrected system $\hat{\bm{H}} \boldsymbol{x} = \hat{\boldsymbol{y}}$, is referred to as the TLS estimate. The left weighting matrix $\bm{\Lambda}$ is typically constructed as a diagonal matrix, \emph{e.g.} $\bm{\Lambda} = \textit{diag}\left[\lambda^{N},...,\lambda^1,\lambda^{0}\right]$, where the diagonal elements represent exponential forgetting factors \cite{rhodeRecursiveGeneralizedTotal2014,arabloueiRecursiveTotalLeastSquares2015,rhodeRecursiveRestrictedTotal2014}. The right weighting matrix $\bm{W}$ is generally chosen as the inverse of the noise covariance matrix to account for measurement uncertainty.

Total least squares (TLS), also known as Orthogonal Regression, is a type of EIV regression, in which both the dependent and independent variables are subject to error. The total least squares approximation is generically equivalent to the best, in the Frobenius norm, low-rank approximation of the data matrix. Hence, TLS is very suitable to our problem in this paper. However, TLS suffers from high computation complexity due to the need for singular value decomposition (SVD) in resolving \eqref{eq:GTLS}, which is not friendly for online implementation with parsimonious onboard computation budgets. It is thus necessary to develop a recursive version.

\subsection{Recursive Total Least-Squares}
Existing literature on solving the TMA problem using TLS \cite{hoAsymptoticallyUnbiasedEstimator2006,vaghefiBearingonlyTargetLocalization2010,guNovelPowerBearingApproach2011,crassidisErrorCovarianceAnalysisTotal2014,xuRobustConstrainedTotal2023,dogancayBearingsonlyTargetLocalization2005,dogancayRelationshipGeometricTranslations2008,farinaTargetTrackingBearings1999} has not adequately addressed the real-time requirements of mobile platforms for target tracking. As the volume of data increases, the computational burden of TLS-based solutions grows substantially. To meet the real-time demands of mobile observers for target state estimation, the algorithm developed in this paper is built based on RTLS \cite{rhodeRecursiveGeneralizedTotal2014}. The RTLS details are presented in Algorithm \ref{alg:RTLS}. Within the RTLS framework, we use $\hat{\boldsymbol{x}}_k$ to represent the estimated TMA parameter vector at time step $k$. According to \eqref{eq:TMA_para}, the moving target's position and velocity can be approximated based on \eqref{eq:p_hat}.
    \begin{equation} \label{eq:p_hat}
        \hat{\boldsymbol{p}}_k = \bm{M}_k \hat{\boldsymbol{x}}_k, \qquad \hat{\boldsymbol{v}}_k = \begin{bmatrix} \bm{0} & \bm{I}  \end{bmatrix} \hat{\boldsymbol{x}}_k
    \end{equation}
\begin{remark}
    In Algorithm \ref{alg:RTLS}, a forgetting factor $0 \ll \lambda < 1$ is introduced, which allows RTLS to prioritize more recent measurements by exponentially weighting them and effectively "forgetting" older information over time \cite{Zhang2022arXiv,10970639}. The introduction of the forgetting factor to enhance the robustness of the RTLS algorithm, and also makes RTLS possibly suitable for tracking time-varying targets. In Line 4 of Algorithm \ref{alg:RTLS}, $\bm{W}_{k}$ should represent the inverse of the covariance matrix of $\boldsymbol{z}_{k+1} = \begin{bmatrix} \tilde{\boldsymbol{h}}_{k+1}^T & \tilde{y}_{k+1}\end{bmatrix}$.However,the cross-covariance between $\tilde{y}_{k}$ and $\tilde{\boldsymbol{h}}_{k}$ is neglected in this implementation, assuming that $\tilde{y}_{k}$ and $\tilde{\boldsymbol{h}}_{k}$ are mutually independent.This independence assumption implies that the bearing measurement noise and observer’s position noise are uncorrelated.While this assumption is not entirely accurate, simulation results demonstrate that the algorithm still performs well compared to the PLKF.
\end{remark}
\vspace{-1em}
\begin{algorithm}
\caption{RTLS Algorithm}  \label{alg:RTLS}
    \begin{algorithmic}[1]
        \State \textbf{Initialization:} $\hat{\boldsymbol{x}}_0 = \bm{0}\in \mathbb{R}^n$ , $\bm{P}_0 = 100\bm{I}\in \mathbb{R}^{n+1 \times n+1}$
        \Function{RTLS}{$\hat{\boldsymbol{x}}_k, \bm{P}_k, \tilde{y}_{k+1}, \tilde{\boldsymbol{h}}_{k+1}, r_{y,k+1}, \bm{R}_{h,k+1}, \lambda$}
        \vspace{0.1cm}
        \State $\boldsymbol{z}_{k+1} = \begin{bmatrix} \tilde{\boldsymbol{h}}_{k+1}^T & \tilde{y}_{k+1} \end{bmatrix}$
        \vspace{0.1cm}
        \State $\bm{W}_{k+1} = \textit{diag} \begin{bmatrix} \bm{R}_{h,k+1}^{-1} & r_{y,k+1}^{-1} \end{bmatrix}$
        \vspace{0.1cm}
        \State $\bm{F}_{k+1}=(\bm{P}_k \boldsymbol{z}_{k + 1}^T)(\lambda + \boldsymbol{z}_{k + 1} \bm{P}_k \boldsymbol{z}_{k + 1}^T)^{-1}$
        \vspace{0.1cm}
        \State $\bm{P}_{k+1}= \lambda^{-1} (\bm{I} - \bm{F}_{k + 1} \boldsymbol{z}_{k + 1})\bm{P}_k$
        \vspace{0.1cm}
        \State $\boldsymbol{v}_{k} = \begin{bmatrix} \hat{\boldsymbol{x}}_k^T & -1 \end{bmatrix}^T$
        \vspace{0.1cm}
        \State $\boldsymbol{v}_{k+1} = \bm{P}_{k+1} (\bm{W}_{k+1} \boldsymbol{v}_{k}) \triangleq [v_{1,k+1}, ...,v_{n+1,k+1}]^T$
        \vspace{0.1cm}
        \State $\hat{\boldsymbol{x}}_{k+1}=-[v_{1,k+1}, ..., v_{n,k+1}]^T/v_{n+1,k+1}$
        \vspace{0.1cm}
        \EndFunction
        \vspace{0.1cm}
        \State \textbf{return:} $\hat{\boldsymbol{x}}_{k+1}, \bm{P}_{k+1}$
    \end{algorithmic}
\end{algorithm}
\vspace{-1em}
\section{Circumnavigation Control}
Due to the absence of distance information, bearing-only TMA requires more restricted observability conditions to ensure the convergence of the state estimation. Theoretical analysis indicates that a mobile observer should exhibit higher-order motion relative to the target, with particular emphasis on motion components that are perpendicular to the bearing direction \cite{ningBearingangleApproachUnknown2024a,liThreeDimensionalBearingOnlyTarget2022}. When a mobile observer follows an orbital trajectory around the target, observability is enhanced, which not only accelerates the convergence of the estimator but also improves the accuracy of parameter estimation \cite{liThreeDimensionalBearingOnlyTarget2022,deghatLocalizationCircumnavigationSlowly2014}. It implies that one must design a proper circumnavigation law for a mobile observer to successfully localize a moving target and persistently track it.

Building on the existing theoretical findings, we propose the following circumnavigation law for a mobile observer to ensure stable behavior.
    \begin{equation} \label{eq:controller}
    \left\{ 
        \begin{aligned}
            &\boldsymbol{u}_{o,k}^{f} = \left( \left \lVert \hat{\boldsymbol{p}}_k - \tilde{\boldsymbol{p}}_{o,k} \right \rVert - \rho \right) \tilde{\boldsymbol{g}}_k \\
            &\boldsymbol{u}_{o,k} = \frac{\min \{U^{f},\| \boldsymbol{u}_{o,k}^{f} \|\}}{\|\boldsymbol{u}_{o,k}^{f}\|} \boldsymbol{u}_{o,k}^{f} + \alpha \tilde{\boldsymbol{g}}^{\perp}_k
        \end{aligned}
    \right.
    \end{equation}
where $ \tilde{\boldsymbol{g}}_k = \begin{bmatrix} \cos \tilde{\theta}_k & \sin \tilde{\theta}_k \end{bmatrix}^T, \tilde{\boldsymbol{g}}_k^{\perp} = \begin{bmatrix} \sin \tilde{\theta}_k & -\cos \tilde{\theta}_k \end{bmatrix}^T$. 
It is evident that $\tilde{\boldsymbol{g}}_k^{\perp}$ is perpendicular to $ \tilde{\boldsymbol{g}}_k$. Here, $\rho$ represents the desired target-observer distance, and $U^{f}$ is the upper bound of the control term $\boldsymbol{u}_{o,k}^{f}$. The parameter $\alpha > 0$ controls the mobile observer's circular maneuverability. The magnitude of $\alpha$ influences the strength of the estimator's observability; increasing $\alpha$ enhances observability. However, the maximum control input constraint limits the value of $\alpha$, such that $U = U^f + \alpha$.
\vspace{-1em}

\section{Numerical Simulations}

In this section, we evaluate the performance of our proposed RTLS algorithm in comparison with the PLKF algorithm \cite{liThreeDimensionalBearingOnlyTarget2022}. The simulation study is conducted in two distinct phases: first, we compare the algorithmic performance under identical parameter conditions; second, we examine how both algorithms respond to varying noise levels while maintaining constant other parameters.



\subsection{Simulation Setup}
To ensure a fair and objective comparison, all algorithms were evaluated under identical parameter conditions. The target's initial position and velocity are given as $\boldsymbol{p}_0 = [10, 5]^T m$ and $\boldsymbol{v}_0 = [1, 1]^T m/s$, respectively. The mobile observer is initialized at $\boldsymbol{p}_{o,0} = [1, 1]^T m$. The forgetting factor is set to $\lambda = 0.999$. For the circumnavigation control law implementation, the key parameters are specified as: $\alpha =5$,$U^{f}=2$, $\rho = 5$.The position estimation error,velocity estimation error and state estimation error are defined as
$e_{\mathrm{p},k} = \|\boldsymbol{p}_{k} - \hat{\boldsymbol{p}}_{k}\|$,
$e_{\mathrm{v},k} = \|\boldsymbol{v}_{k} - \hat{\boldsymbol{v}}_{k}\|$,
$e_{\mathrm{s},k} = \|\boldsymbol{x}_{k} - \hat{\boldsymbol{x}}_{k}\|$.

\subsection{Comparison 1: Accuracy and convergence rate}
In this section, we present a comprehensive comparison of the estimation accuracy and convergence rate between the two algorithms. The simulation parameters are configured as follows: the standard deviation of the mobile observer's position measurement noise is set to $\sigma_{p} = 0.1 m$ , while the standard deviation of the bearing measurement noise is configured as $\sigma_{\theta} = 1^\circ$. 

\begin{figure}[htbp]
        \centering
        \includegraphics[width=\linewidth]{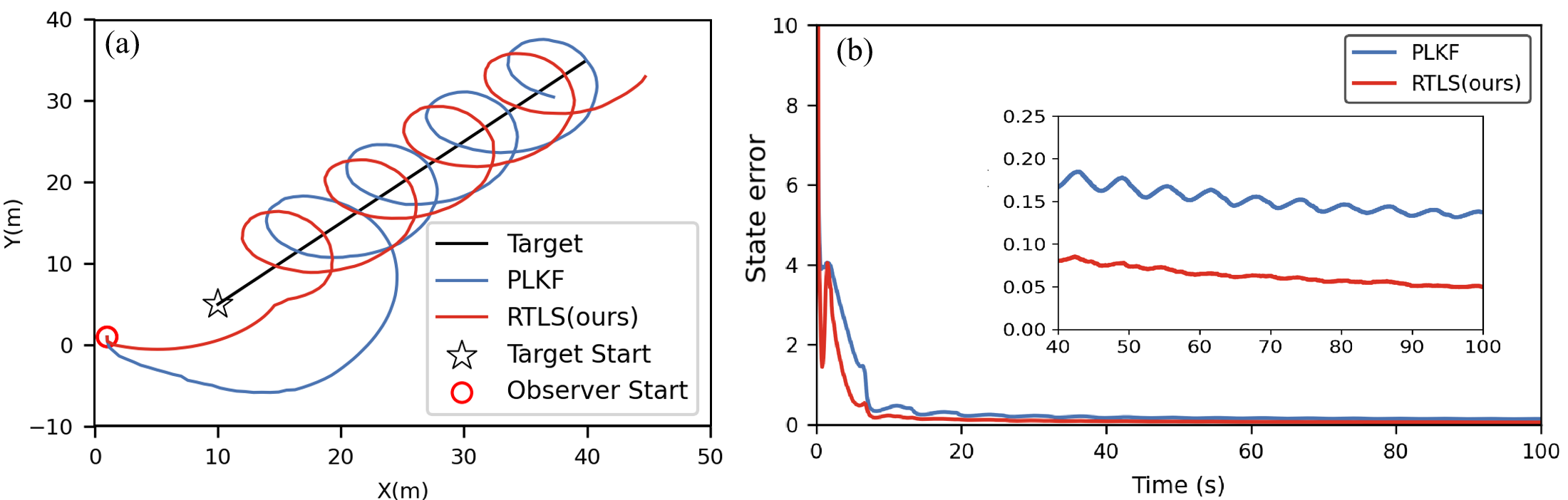}
        \caption{ (a)The simulation trajectories for target circumnavigation.(b)State estimation error of the simulation.}
        \label{fig:tra}
    \end{figure}

\begin{figure}[htbp]
        \centering
        \vspace{-0.3cm} 
        \includegraphics[width=\linewidth]{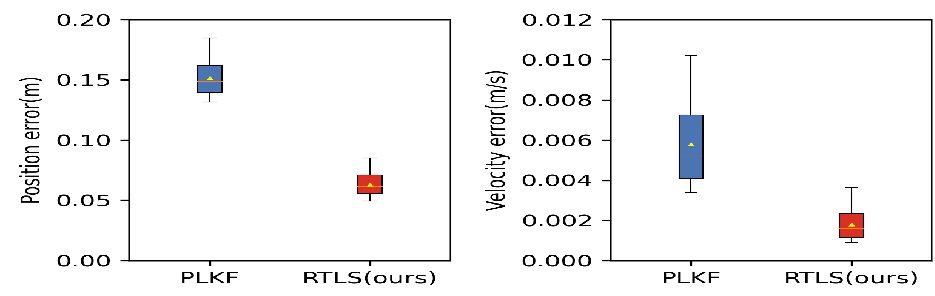}
        \caption{The error distribution of the algorithm. }
        \label{fig:box}
    \end{figure}

Fig.~\ref{fig:tra}(a) illustrates the circumnavigation trajectories of the mobile observer tracking the target using both algorithms. While both algorithms successfully achieve target circumnavigation, our proposed algorithm demonstrates superior performance by reaching the orbital trajectory significantly faster. To ensure statistical reliability, we performed 1000 independent Monte Carlo simulations for each algorithm.The mean state errors across these 1000 trials are presented in Fig.~\ref{fig:tra}(b). The results clearly show that our algorithm achieves substantially lower estimation bias compared to the PLKF algorithm. Furthermore, the analysis of Fig.~\ref{fig:tra} reveals that the RTLS algorithm exhibits faster convergence characteristics. Additionally, the statistical analysis presented in Fig.~\ref{fig:box} demonstrates that our algorithm maintains a more consistent error distribution and enhanced robustness across all simulation trials.
    
    \vspace{-1em} 

    \vspace{-1em} 


\vspace{-1em}
\subsection{Comparison 2: Influence of noise}
In this part, we systematically investigate the individual impacts of bearing measurement noise and observer positioning noise on the performance of both algorithms.

\begin{figure}[htbp]
        \centering
        \vspace{-0.3cm} 
        \includegraphics[width=0.8\linewidth]{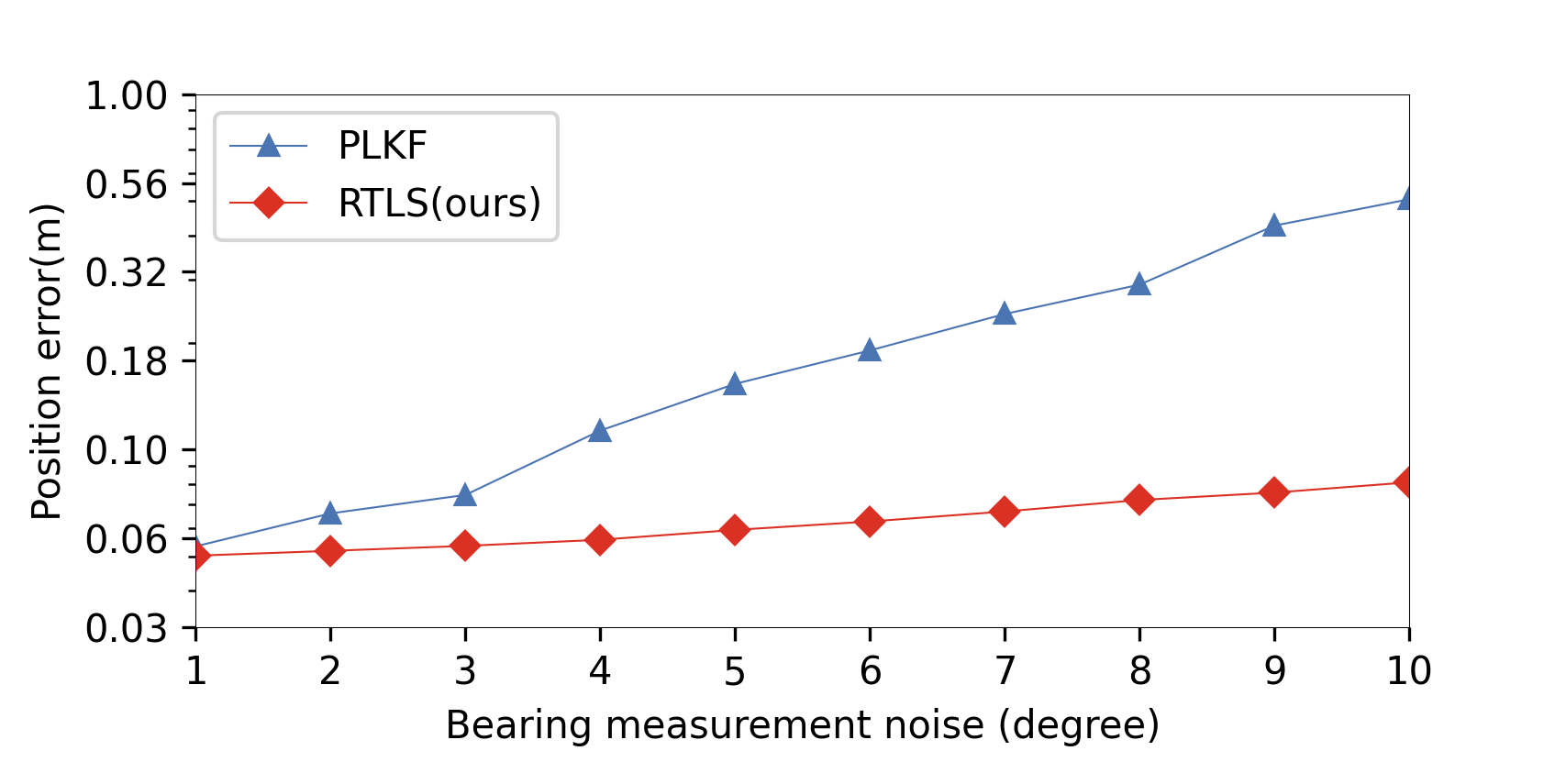}
        \caption{Effect of bearing measurement noise on the error. }
        \label{fig:bearing}
    \end{figure}

\begin{figure}[htbp]
        \centering
        \includegraphics[width=0.8\linewidth]{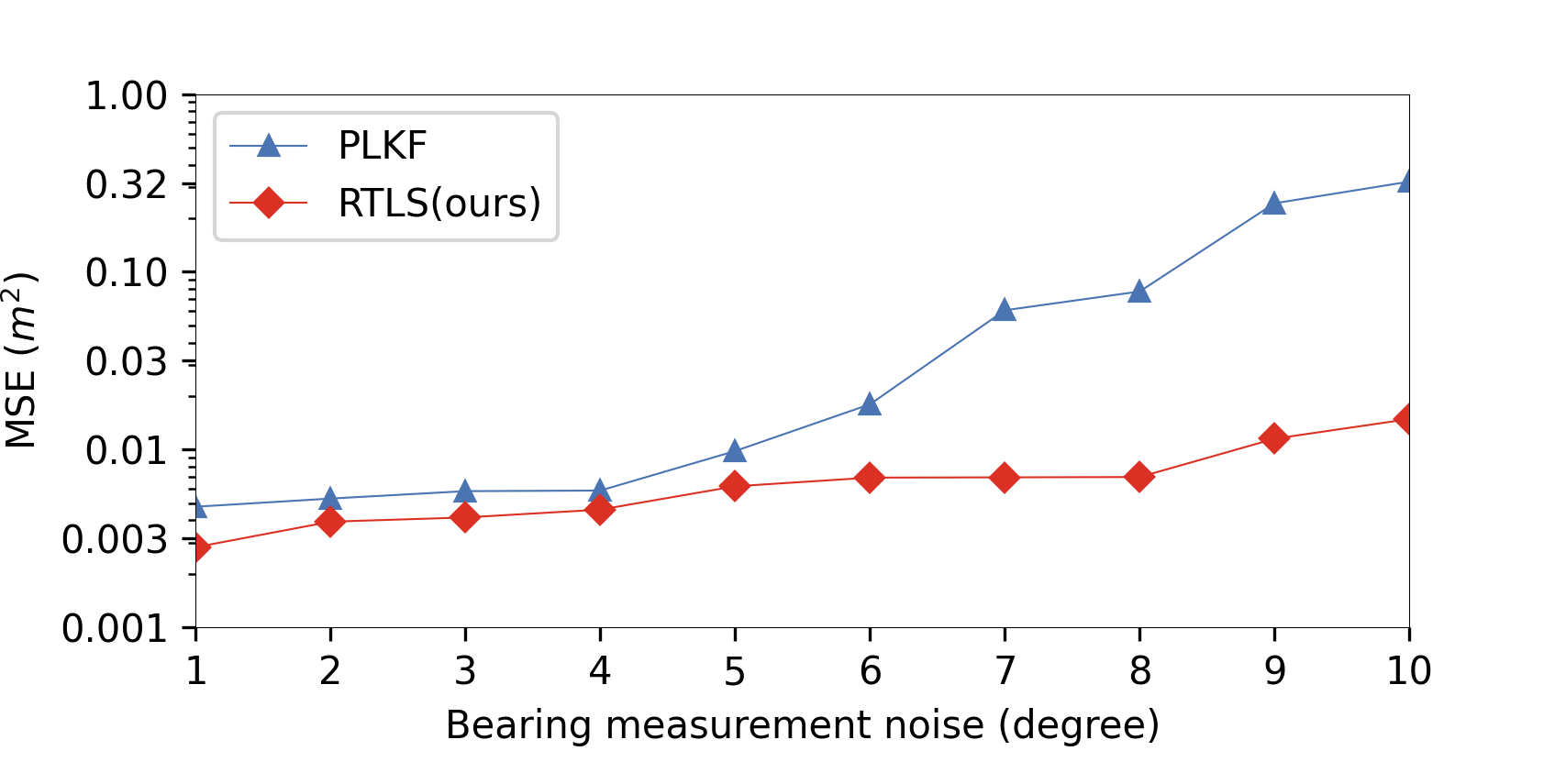}
        \caption{The correlation between the algorithm's MSE and bearing measurement noise. }
        \label{fig:MSEbearing}
    \end{figure}

First, we examine the effect of varying bearing measurement noise while maintaining a constant observer position noise. The observer's position noise standard deviation is fixed at $\sigma_{p} = 1 m$, while the bearing measurement noise standard deviation is increased from $\sigma_{\theta} = 1^\circ$ to $\sigma_{\theta} = 10^\circ$. Fig.~\ref{fig:bearing} demonstrates the influence of bearing measurement noise on position estimation error. As the bearing noise increases, both algorithms exhibit growing position estimation errors; however, the RTLS algorithm demonstrates significantly slower error growth. The corresponding mean square error (MSE) results, presented in Fig.~\ref{fig:MSEbearing}, confirm that the RTLS algorithm consistently maintains lower MSE values compared to the PLKF algorithm across all noise levels.

\begin{figure}[htbp]
        \centering                \vspace{-0.3cm} 
        \includegraphics[width=0.8\linewidth]{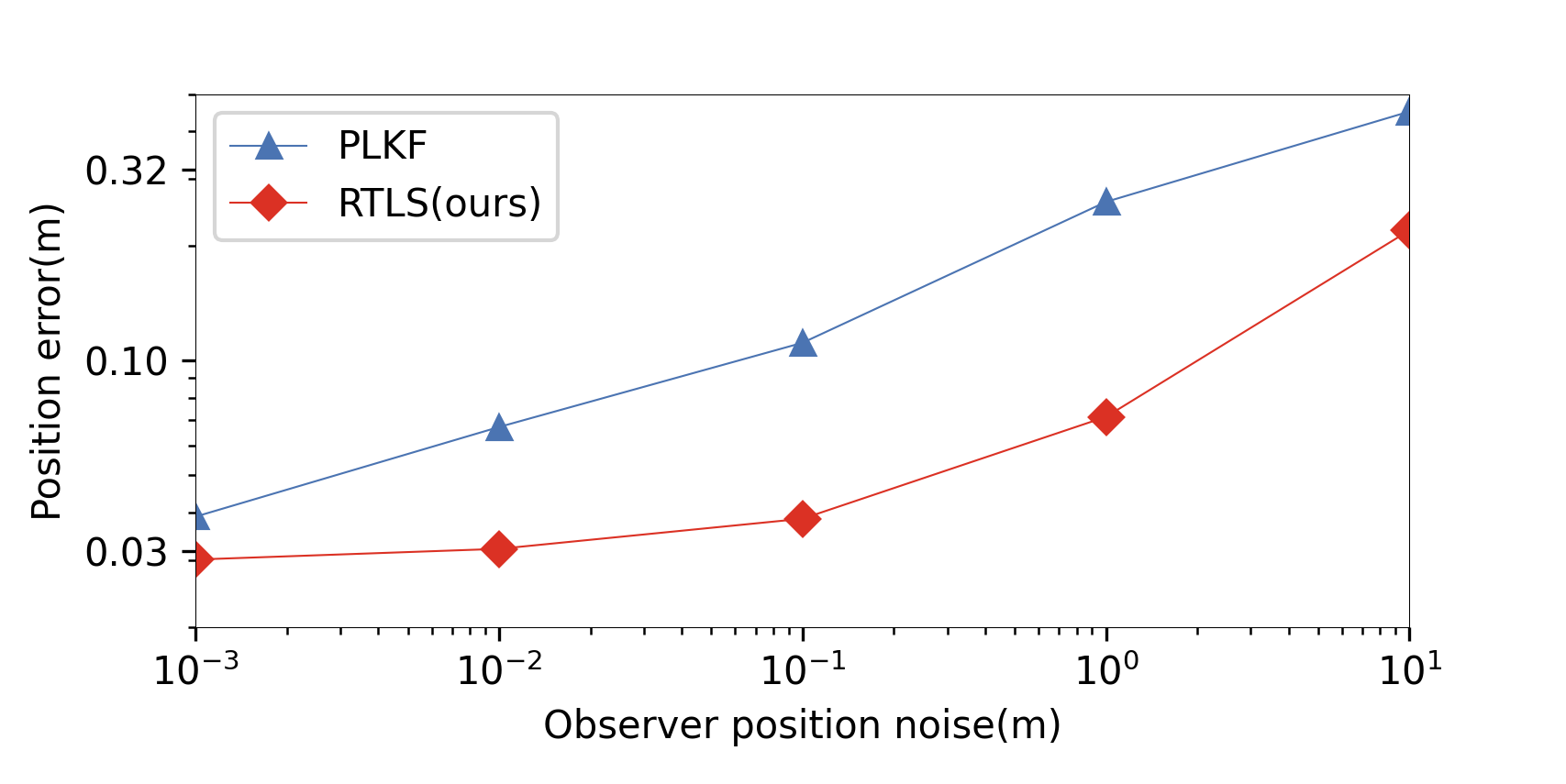}
        \caption{Effect of mobile observer position noise  on the error. }
        \label{fig:pos}
    \end{figure}

\begin{figure}[htbp]
        \centering
        \includegraphics[width=0.8\linewidth]{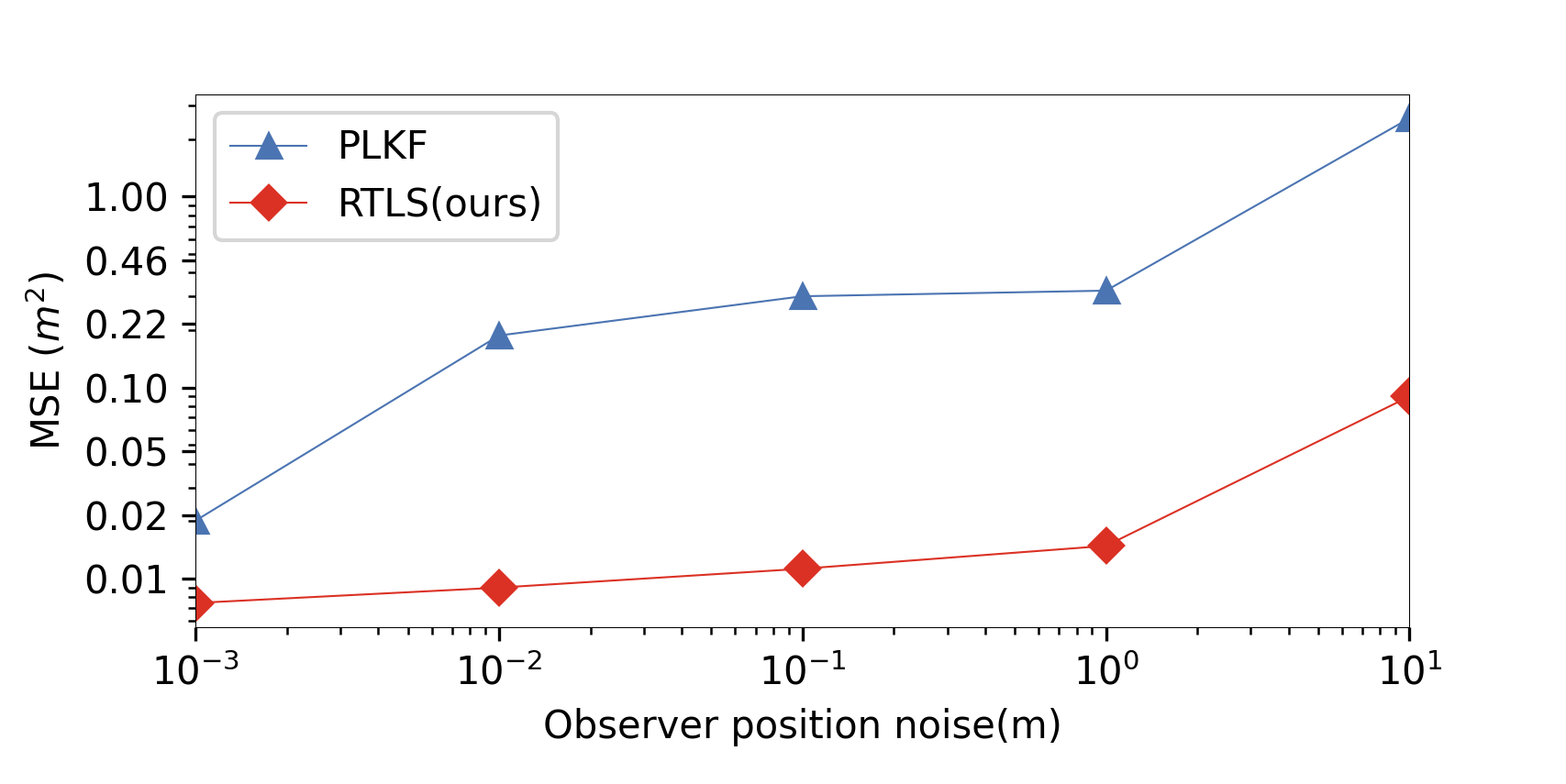}
        \caption{The correlation between the algorithm's MSE and position noise. }
        \label{fig:MSEpos}
    \end{figure}

Subsequently, we analyze the algorithms' performance under increasing observer position errors while maintaining a fixed bearing measurement noise of $\sigma_{\theta} = 5^\circ$. The standard deviation of the observer's position measurement noise is varied from $\sigma_{p} = 0.001 m$ to $\sigma_{p} = 10 m$. The resulting position errors and corresponding MSE values are shown in Fig.~\ref{fig:pos} and Fig.~\ref{fig:MSEpos}, respectively.  While both algorithms experience performance degradation with increasing position noise, the RTLS algorithm maintains superior accuracy, as evidenced by consistently smaller errors and lower MSE values. Furthermore, the RTLS algorithm exhibits more gradual performance degradation, demonstrating enhanced robustness against position measurement noise.

    \vspace{-2em}

\vspace{-2em}
\section{Real-World Experiments}
In this section, we demonstrate the practical efficacy and robustness of the proposed approaches through comprehensive real-world flight experiments.

    \begin{figure}[htbp]
        \centering
        \includegraphics[width=\linewidth]{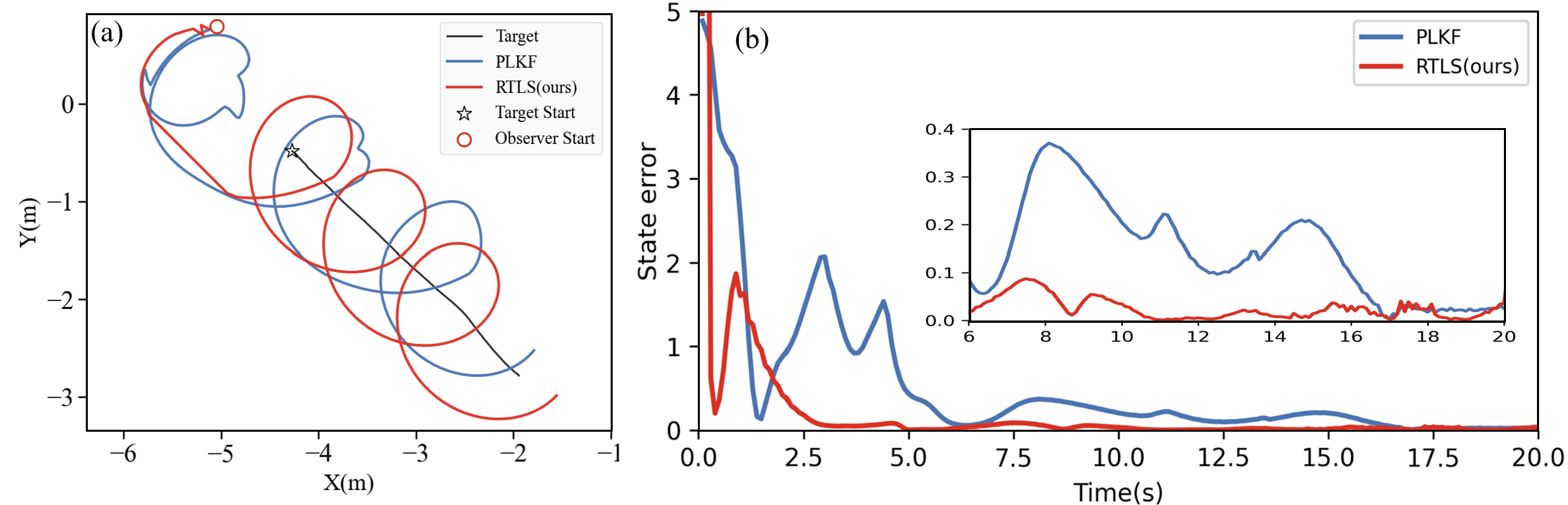}
        \caption{(a) The motion trajectory of the experiment. (b) Position estimation error of the experiment.  }
        \label{fig:exp_tra}
    \end{figure}

In our experimental setup, a Crazyflie quadrotor serves as the aerial observer, tracking a ground-based RoboMaster target (see Fig.~\ref{fig:UAV-TARGET}). The FZMotion optical motion capture system provides precise state measurements for both the quadrotor and the target, which we utilize to calculate bearing information. The RoboMaster was programmed to maintain a constant velocity of $0.3m/s$ along a linear trajectory. For the comparative algorithm evaluation, we maintained identical experimental conditions across all trials. The control parameters were consistently set to $\alpha = 1$,$U^{f}=0.5$, $\rho = 1$. Additionally, both the Crazyflie quadrotor and RoboMaster were initialized at identical starting positions for each experimental run to ensure consistency.

Fig.~\ref{fig:exp_tra}(a) presents the comparative motion trajectories of the Crazyflie quadrotor and RoboMaster target in two representative experimental trials. The corresponding position estimation errors for both algorithms are illustrated in Fig.~\ref{fig:exp_tra}(b). The experimental results demonstrate that the RTLS algorithm achieves significantly faster convergence and maintains superior estimation accuracy compared to the PLKF algorithm, as evidenced by the reduced error metrics.

\section{CONCLUSIONS}
In this paper, we introduced a Recursive Total Least Squares (RTLS) algorithm for bearing-only Target Motion Analysis (TMA) with a mobile observer, effectively addressing both bearing measurement errors and mobile observer positioning errors. The proposed method significantly reduces position estimation bias compared to traditional Pseudo-Linear Kalman Filter (PLKF) methods. Additionally, a circumnavigation controller was designed to enhance system observability and ensure estimator convergence. Extensive simulations and experiments demonstrated the superior accuracy and stability of the RTLS approach, showcasing its effectiveness across various scenarios with different noise conditions. The results underscore the potential of the RTLS algorithm for real-time target localization and tracking applications, offering improved performance over existing techniques.In future work, we aim to integrate real sensor observations into our method to enable implementation in more diverse and practical scenarios, while also considering adaptive models to accommodate more dynamic target behaviors.


\bibliography{2025IROS}
\bibliographystyle{IEEEtran}

\end{document}